\pdfoutput=1

\documentclass[11pt]{article}

\usepackage[final]{acl}

\usepackage{times}
\usepackage{latexsym}

\usepackage[T1]{fontenc}

\usepackage[utf8]{inputenc}

\usepackage{microtype}

\usepackage{inconsolata}

\usepackage{graphicx}

\usepackage{booktabs}
\usepackage{tabularx}
\usepackage{caption}
\usepackage{amsfonts}
\usepackage{amsmath}
\usepackage{color,xcolor} 
\usepackage{xspace}
\usepackage{adjustbox}
\usepackage{multirow}

\newcommand*{\ie}{i.e.\@\xspace}

\usepackage{amssymb}
\usepackage{algpseudocode}
\usepackage{xcolor,colortbl}
\usepackage{algorithm}
\usepackage{dcolumn}
\usepackage{longtable}

\usepackage{graphicx}
\title{Behave Like Experts: Fusing Data with Semantic Knowledge for Effective and Efficient Semantic Map Construction
}

\author{
  Zhu Liu\thanks{These authors contributed equally.} \\
  Shanghai University \\
  \texttt{juniperliuzhu@163.com}
  \And
  Zhen Hu\footnotemark[1], Lei Dai, Ying Liu \\
  Tsinghua University \\
  \texttt{zhenhu317@gmail.com, custdailei@126.com} \\
  \texttt{yingliu@tsinghua.edu.cn}
}

\begin{document}
\maketitle
\begin{abstract}

Semantic map models (SMMs) construct a graph that reflects conceptual similarity by comparing cross-lingual form-meaning mappings. Experts build such models not only by obeying the data constraint—i.e., the connectivity hypothesis, which states that conceptual nodes belonging to the same form must be connected—but also by incorporating their prior semantic knowledge. Current automatic methods ignore this knowledge, which harms both accuracy and efficiency by enlarging the search space of candidate graphs. In this paper, we fuse the distributional semantic knowledge hidden in the internal representations of language models into a top-down, data-driven algorithm. This expert-like behavior re-sorts the candidates in a huge search space, guided by conceptual semantics similarity. Experiments on a self-collected benchmark of 10 datasets and comprehensive metrics show that our method builds SMMs both effectively and efficiently. 

\end{abstract}

\section{Introduction}


Semantic map models (SMMs) build a graph-like conceptual space to reflect the relatedness among concepts or meanings.\footnote{Following the literature in this area, we use the terms \emph{concept}, \emph{meaning}, and \emph{function} interchangeably~\cite{haspelmath2003geometry}.} This relatedness indicates the degree of cross-lingual colexification~\cite{franccois2008semantic}, \ie, how often languages express several meanings with a single form, which may be an affix, a word, a phrase, or even a construction. This theory, originating from linguistic typology~\cite{croft2003typology}, has been applied to several areas, including dialectology~\cite{deng2020semantic}, second language acquisition~\cite{liu2024semantic}, and historical linguistics~\cite{dellert2024causal,narrog2010diachronic}.

\begin{figure}
    \centering
    \includegraphics[width=1.0\linewidth]{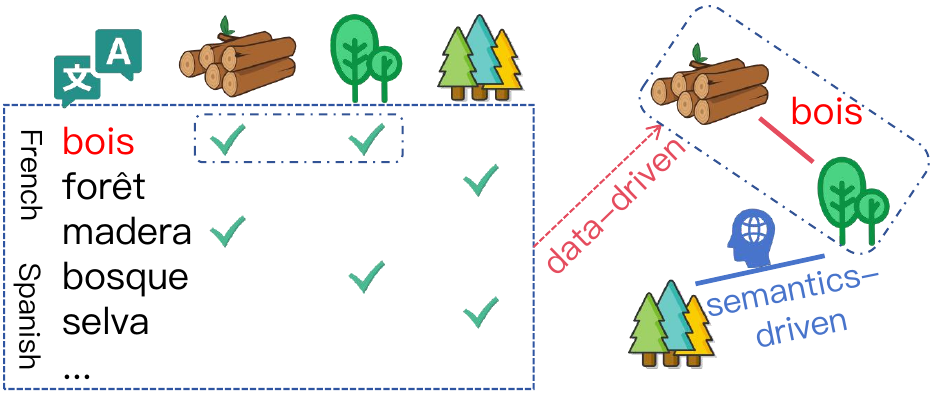}
    \caption{An example of SMM construction over concepts related to \textsc{Tree}.}
    \label{fig:intro}
\end{figure}


Experts consider two aspects when constructing SMMs, as illustrated in Figure~\ref{fig:intro}. The main part is data-driven, obeying constraints inherent to the theory. The data, typically provided as a cross-lingual form-meaning mapping table before construction, indicates which related meanings a form (e.g., a word) has. The constraints~\cite{haspelmath2003geometry} derived from cross-linguistic data are the connectivity hypothesis and the economic principle~\cite{georgakopoulos2018semantic}: each meaning node expressed by one form must be connected (thus topologically close in the graph), and the overall edges should be as sparse as possible. Beyond these data constraints, experts also consider semantic closeness. When several candidate edges are available, they tend to connect concepts with higher semantic similarity, such as metaphor or metonymy analogy~\cite{kovecses2010metaphor}. This information is often rooted in their prior semantic knowledge before observing the cross-lingual data.


Manual construction becomes demanding when scaling to larger data, as it adds or adjusts edges form by form within a large search space. To overcome this limitation of the bottom-up manner, a top-down algorithm~\cite{liu2025top} (dubbed OCMST\footnote{ \underline{\textbf{C}}o-\underline{\textbf{O}}ccurrence-based \underline{\textbf{M}}aximum \underline{\textbf{S}}panning \underline{\textbf{T}}ree.}) based on graph theory, among other automatic methods, formulates the solution as a maximum spanning tree (MST) of a fully connected graph weighted by the co-occurrence count of related nodes. This top-down modeling provides an effective automated way to build SMMs by converting the local data constraint into a global one.

However, this algorithm has several drawbacks when relying on the data constraint alone. First, the number of MSTs is often overwhelmingly large, since many edges share the same integer co-occurrence weight. The algorithm~\cite{liu2025top} prefers chain-like topologies over star-like ones by sorting the standard deviation of degrees in ascending order. However, this additional step costs considerable time due to the large search space. Second, the data alone sometimes cannot reflect the closeness of two specific concepts, especially when the observed data is sparse and noisy, as in low-resource dialect studies. Finally, the non-circle constraint of a tree structure cannot guarantee that every form satisfies the connectivity hypothesis.


To address these drawbacks, we propose OCSMST\footnote{\underline{\textbf{C}}o-\underline{\textbf{O}}ccurrence fused with \underline{\textbf{S}}emantics based \underline{\textbf{M}}aximum \underline{\textbf{S}}panning \underline{\textbf{T}}ree.}, which integrates semantic knowledge into the data-driven top-down algorithm OCMST, inspired by expert construction. Like OCMST, we first build a fully connected graph weighted by meaning co-occurrence counts. Before computing the MST, we incorporate the semantic similarity between every pair of meanings and incorporate it into the edge weights. This similarity can be extracted from language models, ranging from BERT-like encoders to decoder-only large language models, and serves as a prior that regularizes SMM construction. We further propose a post-processing step that merges necessary edges to ensure every form satisfies the connectivity hypothesis. To evaluate SMM performance, we collect 10 datasets spanning diverse domains in linguistic typology and form types, such as function words, content words, and constructions. We also design a comprehensive set of metrics to characterize the generated SMMs. Extensive experiments show that OCSMST builds SMMs more effectively and efficiently than OCMST.


Our contributions are summarized as follows:
\begin{itemize}
\item We propose OCSMST, which integrates the semantic similarity of concepts into OCMST as prior knowledge to guide the search for the best candidate.
\item We introduce a post-processing approach that ensures each form satisfies the connectivity hypothesis.
\item We release a benchmark of 10 datasets spanning diverse domains and grammatical categories, along with a comprehensive set of metrics for evaluating SMMs.
\item Extensive experiments demonstrate the superiority of OCSMST in both effectiveness and efficiency.
\end{itemize}

\section{Related Work}
\subsection{Semantic Map Models}
SMMs represent synchronic or diachronic relationships among meanings by constructing a graph from cross-linguistic comparisons within a semantic domain. The compared forms range from affixes~\cite{cysouw2011expression} and function words~\cite{zhang2017semantic} to content words~\cite{guo2012adjectives, cysouw2007building} and constructions~\cite{malchukov2007ditransitive}. A single form may correspond to multiple meanings depending on context—\emph{multifunctionality} for grammatical items~\cite{haspelmath2003geometry, zhang2015semantic} and \emph{colexification} for lexical items~\cite{franccois2008semantic}. Generally, multifunctionality decreases with larger linguistic units, with function words showing the most context-dependent variation~\cite{zhang2015semantic}.

Experts build the map bottom-up by satisfying \textit{connectivity constraint} and \textit{economic principle} form by form, while applying semantic knowledge to prefer similar concept pairs~\cite{deng2020semantic}. Automatic approaches include the decision-based bottom-up method~\cite{ma2015semantic_en}, network inference~\cite{regier2013inferring}, and the top-down graph-based method OCMST~\cite{liu2025top}. However, all of them are driven solely by cross-lingual data, neglecting semantic knowledge and thus remaining vulnerable to data bias.


\subsection{Conceptual Networks}
Conceptual networks~\cite{gardenfors2000conceptual,gardenfors2014geometry} represent concept relations as a graph, of which SMMs are an instance. The edge weight measuring concept \textbf{similarity} can be constructed in various ways, including co-occurrence~\cite{liu2025top,harvill2022syn2vec}, co-lexification~\cite{franccois2008semantic}, translation~\cite{liu2023crosslingual}, word embeddings~\cite{moullec2024cheaper}, and physical features~\cite{gardenfors2000conceptual}. These measures may correlate with one another~\cite{harvill2022syn2vec, moullec2024cheaper}. Conceptual networks built by models can also reflect human concept perception~\cite{xu2020conceptual,georgakopoulos2022universal}, a topic widely studied in cognitive science.

\section{Preliminary}
\subsection{Semantic Map Models}
Given a semantic domain $\mathcal{D}$, a set of cross-lingual forms $\mathcal{O}$, and their associated functions\footnote{Since SMMs mainly focus on the multiple functions of function words, which are more abstract than the meanings of content words, we adopt the term \textit{function}~\cite{haspelmath2003geometry} to refer to a concept or meaning.} $\mathcal{F}$, the data can be represented as a binary form-function table $T$, indicating which functions each form expresses:
\begin{equation}
    T(i,j)=
    \begin{cases}
    1, & \text{form $i$ has function $j$} \\
    0, & \text{otherwise\footnotemark}.
    \end{cases}
\end{equation}
\footnotetext{This case covers two conditions: form $i$ genuinely lacks function $j$, or function $j$ is simply not observed in the data but may appear elsewhere. Since most datasets fall under the latter, we do not distinguish between them.}



For each form-function pair $(i, j)$ with $T(i, j) = 1$, we collect several evidential examples $\mathcal{E}_{ij}$ in which form $i$ expresses function $j$. This yields a quadruple $(i, j, \mathcal{E}_{ij}, D_j)$, where $D_j$ denotes the definition of function $j$.

From $T$, we infer a graph-like conceptual space $\mathcal{G}$ under the \textit{connectivity hypothesis}, which requires that the functions associated with each form constitute a connected subgraph, reflecting co-lexification~\cite{franccois2008semantic}. The construction also follows the \textit{economic principle}, favoring sparsity by minimizing the number of edges.

The manual approach builds the graph bottom-up, satisfying the connectivity hypothesis form by form. Experts further incorporate the semantic association between two functions as prior knowledge. Although interpretable, this approach suffers from high time complexity due to the large search space and backtracking, limiting its scalability.



\subsection{Top-down Construction}
Alternatively, a \textbf{data-driven} top-down method constructs $G$ globally. Following prior work~\cite{liu2025top}, OCMST \textcolor{black}{builds a co-occurrence graph $G_0^C$} whose nodes are the function set $\mathcal{F}$ and whose edge weights represent how often two functions co-occur in the same form:
\begin{equation}
    G_0^C(j_1, j_2)
    = \left\| T(\cdot, j_1) \odot T(\cdot, j_2) \right\|_1,
\end{equation}
where $T(\cdot, j)$ is the column vector of function $j$, $\odot$ denotes the element-wise AND, and $\|\cdot\|_1$ counts the number of ones in a binary vector.


Then, edge reduction is performed by extracting the maximum spanning tree (MST) from $G_0$, which serves as the optimal conceptual space and transforms local constraints into a global optimization problem. However, when some edges share the same weight, the MST is not unique, and a set of trees $\mathcal{T}_m$ becomes the candidate set. In OCMST, the authors assign the standard deviation of degrees as the weight to each tree $T \in \mathcal{T}_m$, with an inductive bias toward chain-like trees that distribute edge connections more evenly. The top $N$ trees with the maximum weight and the smallest standard deviation of degrees are then presented to experts, who determine the final result.




\section{Method}
\begin{figure*}[t]
    \centering
    \includegraphics[width=1.0\linewidth]{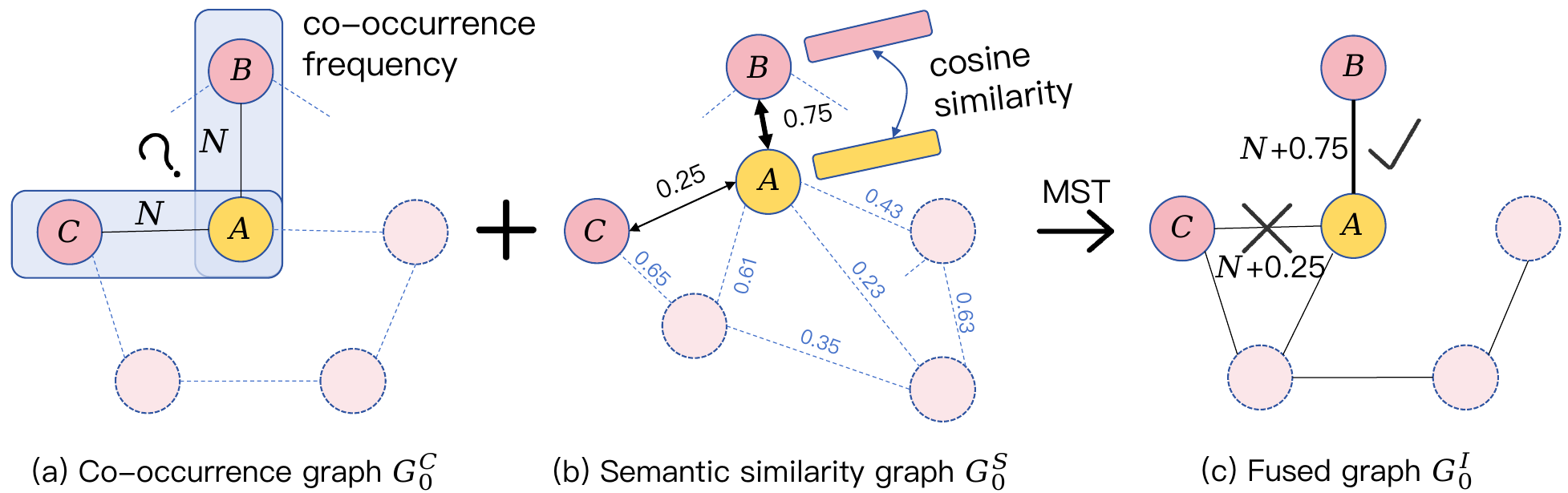}
    \caption{Semantic fusion of OCSMST. Three conceptual nodes, A, B, and C, are highlighted as the focus. In (a), the model is confused about which edge to choose because A--B and A--C have the same weight. In (b), a graph built from semantic similarity forms a semantic space that differentiates the weights of A--B and A--C. The fusion of these two graphs in (c) selects the correct edge.}
    \label{fig:framework}
\end{figure*}

\subsection{$G_0$ with Semantic Knowledge}

The initial graph $G_0$ serves as the starting point of the top-down algorithm and can be regarded as a ``similarity network'' between functions. OCMST measures similarity using cross-lingual co-occurrence frequencies, \ie, from the perspective of colexification. We argue that this typological perspective alone is insufficient to capture concept similarity. We therefore introduce contextual similarity based on distributional semantics: the more similar the contexts in which two concepts appear, the closer they are.

We represent the contextual meaning of function $j$ by an internal representation $e_j$ extracted from a language model. Each function $j$ is associated with a definition $D_j$ and a set of examples $\mathcal{E}_{ij}$, forming a quadruple $(i, j, \mathcal{E}_{ij}, D_j)$ for $i \in \mathcal{F}$. The computation of $e_j$ is detailed in Section~\ref{subsec:LM}. We then construct a contextual similarity graph $G_0^S$, where each edge weight is defined as
\begin{equation}
    w^S_{j_1 j_2} = \mathrm{cosine}(e_{j_1}, e_{j_2}),
\end{equation}
where $\mathrm{cosine}(\cdot, \cdot)$ denotes the cosine similarity between two vectors.

We fuse the co-occurrence graph $G_0^C$ and the contextual similarity graph $G_0^S$, which capture colexification and contextual similarity, respectively. Let $w^C_{j_1 j_2}$ and $w^S_{j_1 j_2}$ denote the edge weights of $G_0^C$ and $G_0^S$. We first normalize the co-occurrence weights:
\begin{equation}
    \tilde{w}^C_{j_1 j_2}
    = \frac{w^C_{j_1 j_2} - \min_{e \in E(G_0^C)} w^C_e}
           {\max_{e \in E(G_0^C)} w^C_e - \min_{e \in E(G_0^C)} w^C_e}.
\end{equation}
The incorporated graph $G_0^I$ then has edge weights
\begin{equation}
    w^I_{j_1 j_2}
    = (1 - \lambda) \cdot \tilde{w}^C_{j_1 j_2}
    + \lambda \cdot w^S_{j_1 j_2},
\label{eqn:combination}
\end{equation}
where $\lambda \in [0, 1]$ balances the two sources. The optimal $\lambda$ is selected based on the average performance across the 10 datasets. Figure~\ref{fig:framework} illustrates the fusion process.



\begin{algorithm}[H]
  \caption{Merge algorithm for full coverage}
  \label{alg:merge}
  \begin{algorithmic}[1]
  \Require A weighted graph $G_0^I = (V, E, w)$, a graph $G$ produced by OCSMST
  \Ensure A graph $G'$ with 100\% coverage
  \State Identify all disconnected components $\mathcal{C}$ by traversing every form
  \State Identify the candidate edges $E_{\text{cand}}$ which belong to $\mathcal{C}$ but not to $G$
  \State $G' \gets G$
  \While{$|\mathcal{C}| > 1$}
    \State Sort $e \in E_{\text{cand}}$ in descending order:

    \State \quad Primary key: number of unconnected components containing edge $e$

    \State \quad Secondary key: weight $w(e)$

    \State Select the top edge $e$ from sorted $E_{\text{cand}}$

    \State Add $e$ to $G'$

    \State Update $\mathcal{C}$

    \State Update $E_{\text{cand}}$

  \EndWhile

  \State \Return $G'$

  \end{algorithmic}

  \end{algorithm}

\subsection{\textcolor{black}{MST with Post-processing}}

After constructing $G_0^I$, we extract its MST following OCMST. Since the fused edge weights are continuous and thus almost all distinct, the optimal tree is unique, allowing us to rank trees directly by their summed weights and avoid the costly standard-deviation computation in OCMST, thereby improving efficiency.

Later, the tree structure without circles is insufficient to meet the connectivity constraint for each form, as a circle is necessary for some forms. To guarantee the 100\% coverage, we develop a post-processing merging algorithm shown in Algorithm~\ref{alg:merge}. Starting from the initial graph with weighted edges, we first identify all disconnected components by traversing every form. For each disconnected form, we add iteratively edges in descending order—first by the number of disconnected components that contain edge $e$, then by weight. After each addition, we update the set of disconnected components and the candidate edge list. This process repeats until the graph becomes fully connected for each form (100\% coverage).

\section{Experiments}
\subsection{SMM Benchmark}


\subsubsection{Datasets}
\label{sec:datasets}

\begin{table*}[ht]
    \centering
    \small
    \setlength{\tabcolsep}{4pt}
    \begin{tabular}{cl|l|cccc|ccc}
        \toprule
        Id & Name (Domain) & Source & \#L & \#F & \#Fc & Sparsity & Avg\_D & Div\_D & \#E \\
        \midrule
        \multicolumn{9}{l}{\textit{Function words}} \\
        1 & SUPPLEMENTARY adv. & \cite{guo2010semantic} & 37 & 82 & 18 & 0.816 & 2.111 & 1.560 & 19\\
        2 & PERFECTIVE marker \textit{De} & \cite{liu2015semantic_de} & 14 & 14 & 9 & 0.737  & 2.00 & 1.25 & 9 \\
        3 & AND \textit{Hé} & \cite{ma2015semantic_zhai} & 55 & 55 & 29 & 0.801 & 2.000 & 1.509 & 29 \\
        4 & FINALLY \textit{Dàodǐ} & \cite{yu2023synonyms} & - & 31 & 8 & 0.605 & 2.25 & 1.56& 9 \\
        5 & QUANTIFIER & \cite{li2011cross} & 90 & 90 & 14 & 0.590 & 1.857 & 1.059 & 13 \\
        \midrule

        \multicolumn{9}{l}{\textit{Content words}} \\
        6 & EAT verbs & \cite{deng2020semantic} & 42 & 42 & 17 & 0.648 & 1.882 & 1.711 & 16 \\
        7 & MAKE \textit{Zuò} & \cite{liu2024semantic} & 17 & 17 & 8 & 0.375 & 1.750 & 1.090 & 7\\
        8 & HIT \textit{Dǎ} & \cite{liu2024semantic} & 14 & 14 & 19 & 0.417 & 1.895 & 2.292 & 18 \\
        9 & COME \textit{Laí} & \cite{liu2024semantic} & 14 & 14 & 8 & 0.420 & 1.750 & 1.392 & 7\\
        \midrule

        \multicolumn{9}{l}{\textit{Constructions}} \\
        10 & Ditransitive & \cite{guo2013ditransitive} & 6 & 29 & 34 & 0.737 & 1.941 & 1.454 & 33 \\
    \bottomrule
    \end{tabular}
    \caption{Well-established cases from typological literature. Three grammatical categories are split into three parts. The core concept in each case is capitalized with the core Chinese word in italics. L, F, Fc, E, Avg\_D and Div\_D represent languages, forms, functions, average of degrees, standard derivation of degrees.}
    \label{tab:cases_info}
\end{table*}

We collect 10 datasets from the typological literature, covering diverse domains, grammatical categories, and scales. Detailed information is listed in Table~\ref{tab:cases_info}. The selection process is as follows. First, we identify classical or recent papers on SMMs, mainly from linguistic typology. Second, we extract the form-function table $T$, together with the definition $D_j$ of each function and the examples $\mathcal{E}_{ij}$ for each form-function pair. The ground truth (GT) annotated by linguistic experts is also collected as a reference for evaluating the automatic algorithm.

We then report several statistics for each dataset. The number of languages (\#L), forms (\#F), functions (\#Fc), and the sparsity of the form-function table (the proportion of zeros) reflect the dataset scale, while the average degree (Avg\_D), the standard deviation of degrees (Div\_D), and the number of edges (\#E) characterize the topology of the ground truth (GT). Note that we extend the SUPPLEMENT adverbs dataset used as a case study in OCMST~\cite{liu2025top} by adding more forms across more languages and dialects. Manual construction becomes increasingly difficult at larger scales, which is unavoidable as more languages are observed. The GT topology exhibits a chain-like structure with an average degree of around two and a number of edges close to the number of function nodes, as also observed in OCMST.





\subsubsection{Metrics}

We design a comprehensive set of metrics to evaluate the graph $G'$ generated by the algorithm, as listed in Table~\ref{tab:description}. The detailed equations for each metric are given in Appendix~\ref{app:metrics}.

\textbf{Intrinsic metrics} measure the graph itself, involving network topology (Div\_D) and the ability to represent observed data (Cov). A high coverage (Cov) indicates that each observed form has a connected area in the SMM, as required by the connectivity hypothesis. A low Div\_D indicates a chain-like topology, which is preferred in SMMs because it enhances entailment predictability, as verified in OCMST~\cite{liu2025top}.

\textbf{Extrinsic metrics}, by contrast, evaluate graphs against the reference GT. Accuracy (Acc) targets the adjacency matrix of the graph, while precision (Prec), recall (Rec), and F1 focus on edge matching. Higher values are better for all of them.

\begin{table}[h]
    \centering
    \small
    \begin{tabularx}{0.99\linewidth}{cXc} 
    \toprule
        \textbf{Metric} & \textbf{Description} & \textbf{Trend} \\
    \midrule
        \multicolumn{3}{l}{\textit{Intrinsic}} \\
        Div\_D & Standard deviation of node degrees & $\downarrow$ \\
        Cov & Proportion of covered instances & $\uparrow$ \\
    \midrule
        \multicolumn{3}{l}{\textit{Extrinsic}} \\
        Acc & \textcolor{black}{Matched / total elements} in $\mathcal{G}$ & $\uparrow$ \\
        Prec & Matched edges / generated edges & $\uparrow$ \\
        Rec & Matched edges / annotated edges & $\uparrow$ \\
        F1 & Harmonic mean of Prec and Rec & $\uparrow$ \\
    \bottomrule
    \end{tabularx}
    \caption{Metrics for evaluating SMMs. Arrows indicate the preferred direction.}
    \label{tab:description}
\end{table}


Although extrinsic metrics imply that edge selection can be viewed as a classical \textbf{classification} task, we also treat it as a \textbf{retrieval} task. This is because MST-based algorithms produce a set of candidate graphs ranked by the overall weight or Div\_D, as in OCMST. We can therefore evaluate the top-$N$ candidates by testing either the best one (denoted as METRIC@N$^*$) or the averaged one (denoted as METRIC@N). This rationale also applies to the human-in-the-loop process~\cite{liu2025xism}, where multiple plausible results are presented to experts for final selection.

\subsection{Extracting Internal Representations from Language Models}
\label{subsec:LM}

We extract internal representations using a diverse set of multilingual models, spanning both encoder-only and decoder-only architectures, under various extraction strategies.

\paragraph{Encoder-only models.}
\textsc{bge-m3}\footnote{\url{https://huggingface.co/BAAI/bge-m3}} (569M) is a multilingual embedding model supporting dense, sparse, and multi-vector retrieval across 170+ languages.
\textsc{embeddinggemma-300m}\footnote{\url{https://huggingface.co/google/embeddinggemma-300m}} (308M) is built on Gemma-3\footnote{\url{https://huggingface.co/docs/transformers/model_doc/gemma3}} with flexible embedding dimensions, covering 100+ languages.
\textsc{mmBERT-base}\footnote{\url{https://huggingface.co/jhu-clsp/mmBERT-base}} (307M), based on ModernBERT\footnote{\url{https://huggingface.co/answerdotai/ModernBERT-base}}, is pretrained on 1,833 languages but \emph{not} finetuned for semantic matching.

\paragraph{Decoder-only LLMs.}
\textsc{Qwen3-Embedding}\footnote{\url{https://huggingface.co/collections/Qwen/qwen3-embedding}} is a series of text embedding models built on the dense Qwen3 foundation models, supporting a 32K context window and over 100 languages across three parameter scales (0.6B, 4B, and 8B).

\paragraph{Embedding Extraction Strategies}
We use four strategies to aggregate the available information for each function and extract its internal representation.
\textbf{Def-only} uses only the definitional sentence.
\textbf{Exm-only} uses only the examples, indexed by the target word form.
\textbf{DefExm-sim} simply concatenates the definition and examples.
\textbf{DefExm-com} organizes them in a structured template: ``The domain is about <$A$>. The explanation of the function is <$D$>, along with the following examples: <$\mathcal{E}$>'', where $A$, $D$, and $\mathcal{E}$ denote the domain description, definition, and examples, respectively. Detailed descriptions can be found in Table~\ref{tab:aggregation-strategies}.


\begin{table}[t]
\centering
\small
\begin{tabularx}{\columnwidth}{l X}
\toprule
\textbf{Strategy} & \textbf{Input and extraction} \\
\midrule
Def-only & Definition $D$ only; sentence embedding. \\
\addlinespace
Exm-only & Examples $\mathcal{E}$ only, indexed by target word form; last hidden state at target token (averaged if split). \\
\addlinespace
DefExm-sim & Concatenation of $D$ and $\mathcal{E}$; sentence embedding. \\
\addlinespace
DefExm-com & Structured combination of domain $A$, definition $D$, and examples $\mathcal{E}$; sentence embedding. \\
\bottomrule
\end{tabularx}

\caption{\textcolor{black}{Four information aggregation strategies.}}
\label{tab:aggregation-strategies}
\end{table}

\subsection{Baselines}
We compare our method OCSMST with several baselines. OCMST~\cite{liu2025top} uses only co-occurrence frequency as edge weights. We also build two variants of OCMST: OCMST\_J and OCMST\_D replace the raw co-occurrence count with the Jaccard and Dice similarity, respectively, regularizing OCMST through set unions, as described in Appendix~\ref{app:variants}.



Furthermore, we employ two strong LLM-based agents, Qwen\footnote{\url{https://www.qianwen.com/}} and Codex, to generate SMMs end-to-end. Powered by Qwen3-Max~\cite{yang2025qwen3technicalreport} and GPT-5.6-Luna\footnote{\url{https://developers.openai.com/api/docs/models/gpt-5.6-luna}} respectively, both follow tailored instructions but adopt distinct strategies. Qwen uses a coverage-based greedy heuristic that prioritizes function pairs co-occurring in the most form records and adds edges to repair disconnected subgraphs. Codex, by contrast, formulates SMM construction as a minimum-cardinality hitting-set problem: it restricts candidates to function pairs co-occurring in at least one record and searches for the smallest edge set that keeps the functions in every record connected. Details are in Appendix~\ref{app:skill}.

\section{Results}
\begin{table*}[]
    \centering
    \small
    \begin{tabular}{c|cc|cccc|cccc|c}
    \toprule
     Method  & Cov & Div\_D$\downarrow$ & Acc & Prec & Rec & F1 & F1@5 & F1 @5$^*$ & F1@10 & F1@10$^*$ & Time$\downarrow$\\
     \midrule
      OCMST   & \underline{97.96} & \underline{1.43} & 88.83 & 62.31 & 58.65 & 60.31 & 54.68 & 65.31 & 52.62 & 66.84 & 14.53 \\
      OCMST\_J & 74.10 & \textbf{0.84} & 82.87 & 43.67 & 41.05 & 42.25 & 44.06 & 48.86 & 43.18 & 50.13 & 0.22 \\
      OCMST\_D & 74.10 & \textbf{0.84} & 82.84 & 43.37 & 40.75 & 41.94 & 44.45 & 49.21 & 44.43 & 52.81 & 2.60 \\
      Qwen & 86.81 & 2.04 & 81.02 & 40.82 & 45.51 & 42.49 & - & - & - & - & 357.4 \\
      Codex & \textbf{100} & 2.54 & 87.95 & 56.73 & 56.19 & 56.44 & - & - & - & - & 490 \\
      \midrule 
      OCSMST & 97.59 & 1.43 & \underline{92.25} & \underline{71.84} & \underline{68.01} & \underline{69.76} & \underline{66.88} & \underline{74.69} & \underline{66.11} & \underline{75.81} & \textbf{0.09}\\
      OCSMST\_m & \textbf{100} & 1.53 & \textbf{93.02} & \textbf{72.25} & \textbf{71.94} & \textbf{72.07} & \textbf{69.20} & \textbf{76.83} & \textbf{68.56} & \textbf{77.94} & \underline{0.14} \\
      \bottomrule
    \end{tabular}
    \caption{Overall performance for baselines and our method OCSMST along with its post-processing version (OCSMST w. m). Except Div\_D and Time with seconds, other metrics are shown the percent.}
    \label{tab:performance}
\end{table*}


We compare our proposed methods, OCSMST and OCSMST\_m (OCSMST with post-processing that merges edges for unconnected forms), against other methods in Table~\ref{tab:performance}. We set the fusion weight of semantic embeddings to $\lambda = 0.1$ and adopt Qwen3-Embedding 8B with the Def-only strategy, based on overall dataset performance (see Subsection~\ref{sec:lambda} and Subsection~\ref{sec:semantic fusion}).


OCSMST achieves strong performance on extrinsic metrics under both the classification and retrieval formulations, improving F1 by nearly 10 points over OCMST (from 60.31 to 69.76) without semantic fusion. This highlights the importance of prior semantic knowledge in guiding SMM construction. Due to the tree structure, coverage (Cov) cannot reach 100\%. Our post-processing, which merges edges to the minimal set, connects all forms in the graph and yields better performance on other extrinsic metrics. For the two OCMST variants, OCMST\_J and OCMST\_D, performance degrades significantly, with nearly a 20\% drop in F1 and lower coverage, suggesting that edge weights are better not regularized by set union size. They do, however, prefer a chain-like topology with low Div\_D.

The LLM-agent baselines generally underperform OCSMST. Qwen performs worse, whereas Codex achieves somewhat better results with connectivity ensured. This difference likely stems from differences in foundation model capabilities, which lead to different algorithmic choices. Although both agents capture key structural principles such as connectivity and economy, they have limited understanding of the semantic-map objective and do not fully leverage relevant semantic information or linguistic priors. Consequently, their edge-selection decisions deviate from the expert construction process, which integrates structural constraints with fine-grained semantic evidence.



We further present dataset-specific comparisons between OCMST and OCSMST\_m in Appendix~\ref{app:case_f1}. Performance improves in 7 out of 10 cases, with four datasets (Datasets 1, 4, 7, and 9) reaching an F1 of 100\%. Notably, on Dataset 1, the SUPPLEMENT adverbs case study in OCMST~\cite{liu2025top}, our method generates an SMM that exactly matches human experts, demonstrating its competitiveness.

For \textbf{efficiency}, OCSMST and its merged version are the fastest, comparable to OCMST\_J and OCMST\_D, since all three make the initial edge weights continuous and thus mostly distinct. OCSMST achieves a more thorough discrete-to-continuous transition through semantic embeddings, whereas OCMST\_J and OCMST\_D merely divide by the integer union size. OCMST costs over 100 times as much, as thousands or more candidate trees, produced by an initial graph with many repeated edge weights, must be sorted. Qwen and Codex require several minutes per SMM, far exceeding the MST-based baselines, because their agent-based pipelines add iterative LLM inference, tool invocation, and intermediate-state management, while their algorithms incur extra overhead from repeated candidate evaluation and connectivity checks (Qwen) or recursive search over edge combinations (Codex).




\subsection{Different Settings of Semantic Fusion}
\label{sec:semantic fusion}

\begin{table}[t]
    \centering
    \small
    \setlength{\tabcolsep}{8pt}
    \renewcommand{\arraystretch}{1.15}
    \begin{tabular}{l l c}
    \toprule
    \textbf{Model} & \textbf{Type} & \textbf{F1} \\
    \midrule
    OCMST          & --            & 60.31 \\
    \midrule
    \multirow{4}{*}{Qwen3-Embedding 0.6B}
        & Def-only      & 64.13 \\
        & Exm-only      & 54.09 \\
        & DefExm-sim    & \textbf{64.79} \\
        & DefExm-com    & 59.31 \\
    \midrule
    \multirow{4}{*}{Qwen3-Embedding 4B}
        & Def-only      & 67.51 \\
        & Exm-only      & 54.76 \\
        & DefExm-sim    & \textbf{70.51} \\
        & DefExm-com    & 67.33 \\
    \midrule
    \multirow{4}{*}{Qwen3-Embedding 8B}
        & Def-only      & \textbf{69.76} \\
        & Exm-only      & 56.38 \\
        & DefExm-sim    & 68.89 \\
        & DefExm-com    & 68.21 \\
    \midrule
    \multirow{4}{*}{bge-m3}
        & Def-only      & \textbf{64.95} \\
        & Exm-only      & 56.43 \\
        & DefExm-sim    & 62.72 \\
        & DefExm-com    & 60.31 \\
    \midrule
    \multirow{4}{*}{embeddinggemma-300m}
        & Def-only      & 63.94 \\
        & Exm-only      & 58.49 \\
        & DefExm-sim    & 62.85 \\
        & DefExm-com    & \textbf{65.08} \\
    \midrule
    \multirow{4}{*}{mmBERT-base}
        & Def-only      & \textbf{63.39} \\
        & Exm-only      & 60.12 \\
        & DefExm-sim    & 58.41 \\
        & DefExm-com    & 59.96 \\
    \bottomrule
    \end{tabular}
    \caption{Average Top-1 $F_1$ (\%) over 10 cases under different settings of semantic fusion with $\lambda=0.1$.}
    \label{tab:semantic_fusion}
\end{table}

We illustrate the F1 scores for different semantic fusion settings in Table~\ref{tab:semantic_fusion}, covering several models and four fusion types, as described in Section~\ref{subsec:LM}. All models leveraging the optimal strategy outperform OCMST, and Qwen3 series LLMs perform better than models with fewer parameters. The worst model is mmBERT-base, expected since it is not finetuned on semantic matching tasks. No single fusion type is consistently best across models, though three models perform best with \textbf{Def-only}, suggesting that the definition alone provides sufficient semantic knowledge. Notably, minimal information works best, even outperforming the combination of definition and examples, which is practically advantageous since examples are not always available for each form-function pair.

\subsection{Tradeoff Between Co-occurrence and Semantics}
\label{sec:lambda}

\begin{figure}[h!]
    \centering
    \includegraphics[width=1.0\linewidth]{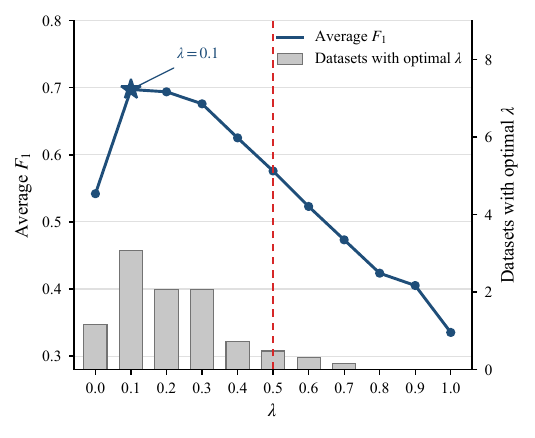}
    \caption{Average F1 score and the number of datasets achieving the optimal $\lambda$. The tradeoff is indicated by the red dotted line, and the optimal $\lambda$ is marked with a star.}
    \label{fig:optimal_lambda}
\end{figure}

We incorporate semantic embeddings into the co-occurrence graph $G_0$ with a weight $\lambda$ that balances co-occurrence and semantic knowledge, as shown in Eqn.~\ref{eqn:combination}. To examine how these two sources interact, we run all 10 datasets and compute the F1 score as $\lambda$ varies from 0 to 1, along with the number of datasets achieving each optimal $\lambda$. Figure~\ref{fig:optimal_lambda} shows the results. Performance declines significantly once $\lambda > 0.5$, where co-occurrence and semantic similarity are weighted equally, indicating that semantics serves as a supplementary feature—expected, given that SMMs originate from linguistic typology. Notably, performance nearly collapses when only semantics is used ($\lambda = 1.0$), suggesting that the association between functions is not equivalent to semantic similarity. Based on this trend, we set $\lambda = 0.1$ in our experiments.

\subsection{Role of Semantic Fusion: A Case Study}

\begin{figure}[h!]
    \centering
    \includegraphics[width=0.7\linewidth]{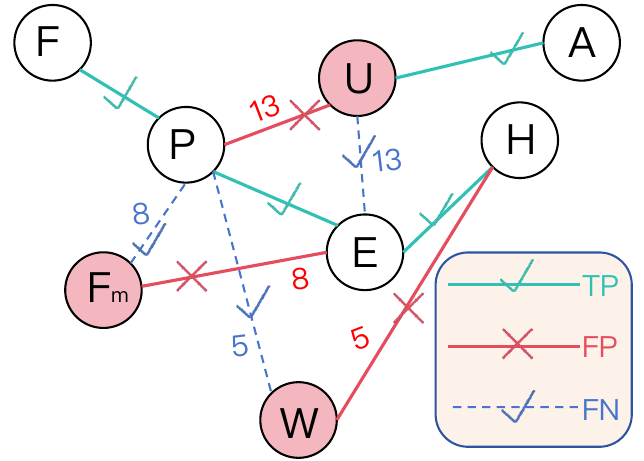}
    \caption{SMMs for Dataset 9: OCMST vs.\ OCSMST, where OCSMST coincides with the GT. Semantic fusion identifies the correct edges for nodes W (WRITE), Fm (FORM), and U (UNDERTAKE).}
    \label{fig:SMM_case_study}
\end{figure}

To illustrate the role of semantic fusion, we examine the 7th dataset on the semantic domain of MAKE. Its form-function table is provided in Appendix~\ref{app:city_features}, where function abbreviations are given in the header and the languages are Chinese dialects from different cities. Figure~\ref{fig:SMM_case_study} compares the semantic maps generated by OCMST and OCSMST against the GT, where colors and markers indicate three types of connection errors. Here, OCSMST produces an SMM identical to the GT. True positives (TP, green lines) are edges produced by both OCMST and the GT/OCSMST; false positives (FP, red lines) are edges only from OCMST; false negatives (FN, blue dotted lines) are edges only from the GT/OCSMST. Edge weights denote the co-occurrence frequency of the two functions.

Semantic fusion clearly helps select the correct edge when plausible edges around a node have similar weights. For example, the function WRITE has two edges with the same weight of 5: W--P (PRODUCE) and W--H (HOLD), implying that co-occurrence alone is insufficient. Guided by semantics—where WRITE is closer to PRODUCE than to HOLD—OCSMST correctly selects W--P, consistent with expert annotation. Similar patterns appear in other FPs and FNs, such as Fm (FORM)--P over Fm--E (ENGAGE) and U (UNDERTAKE)--E over U--P.

\section{Conclusion}
In this paper, we propose OCSMST, which incorporates semantic knowledge into the co-occurrence frequency used by OCMST for building semantic maps, inspired by typological experts. OCSMST, together with its post-processing step that merges necessary edges to satisfy the connectivity hypothesis for each form, achieves significant gains in both effectiveness and efficiency on a well-designed benchmark we construct. This makes SMMs promising for practical applications such as foreign language teaching.


\section*{Limitations}
We acknowledge several limitations. First, although our datasets cover a wide range of topics, they remain limited in language coverage, literature sources (mostly Chinese papers), and data scale. Second, our study relies on a series of language models and may thus be biased toward specific ones. Third, we do not include evaluations from linguistic experts on SMMs, which may affect the actual performance and potential applications.

\section*{Ethics Statement}
This study does not involve any ethical issues.


\bibliography{custom}

\appendix

\section{Appendix}
\algnewcommand{\Sort}[1]{\State \textbf{sort} #1}

\subsection{Evaluation Metrics}
\label{app:metrics}

We define metrics to evaluate the conceptual space $G$ both intrinsically and extrinsically.

\subsubsection{Intrinsic Metrics}

\begin{itemize}
    \item \textbf{Coverage}: proportion of connected forms $F'$ among observed forms $F$: 
    \[
    \text{Coverage} = \frac{|F'|}{|F|}
    \]
    \item \textbf{Div\_D}: standard deviation of node degrees: 
    \[
    \text{Div\_D} = \sqrt{\frac{1}{|V|} \sum_{v \in V} (\deg(v) - \text{Avg\_D})^2}
    \]
\end{itemize}

\subsubsection{Extrinsic Metrics}

Let $G(w>0)$ denote edges with positive weight.  

\begin{itemize}
    \item \textbf{Accuracy (Acc)}: fraction of correctly predicted edges: 
    \[
    \text{Acc} = \frac{|G \cap G^\ast|}{|G^\ast|}
    \]
    \item \textbf{Precision (Prec)}: correctly predicted edges over predicted edges: 
    \[
    \text{Prec} = \frac{|G(w>0) \cap G^\ast(w>0)|}{|G(w>0)|}
    \]
    \item \textbf{Recall (Rec)}: correctly retrieved edges over gold edges: 
    \[
    \text{Rec} = \frac{|G(w>0) \cap G^\ast(w>0)|}{|G^\ast(w>0)|}
    \]
    \item \textbf{F1-score (F1)}: harmonic mean of Precision and Recall: 
    \[
    \text{F1} = \frac{2 \cdot \text{Prec} \cdot \text{Rec}}{\text{Prec} + \text{Rec}}
    \]
\end{itemize}

\subsection{Variants of OCMST}
\label{app:variants}

We introduce two variants of OCMST that replace the raw co-occurrence count in the initial graph $G_0^C$ with set-based similarity measures.

Let $T(\cdot, j)$ denote the binary column vector of function $j$, where $T(i, j) = 1$ if form $i$ expresses function $j$, and $T(i, j) = 0$ otherwise. For two functions $j_1$ and $j_2$, let
\[
    \mathcal{S}_{j_1} = \{i \mid T(i, j_1) = 1\}, \quad
    \mathcal{S}_{j_2} = \{i \mid T(i, j_2) = 1\}
\]
denote the sets of forms that express $j_1$ and $j_2$, respectively.

\subsubsection{OCMST\_J}

OCMST\_J replaces the co-occurrence count with the Jaccard similarity between $\mathcal{S}_{j_1}$ and $\mathcal{S}_{j_2}$:
\[
    G_0^J(j_1, j_2)
    = \frac{|\mathcal{S}_{j_1} \cap \mathcal{S}_{j_2}|}{|\mathcal{S}_{j_1} \cup \mathcal{S}_{j_2}|}.
\]

\subsubsection{OCMST\_D}

OCMST\_D replaces the co-occurrence count with the Dice similarity between $\mathcal{S}_{j_1}$ and $\mathcal{S}_{j_2}$:
\[
    G_0^D(j_1, j_2)
    = \frac{2 \cdot |\mathcal{S}_{j_1} \cap \mathcal{S}_{j_2}|}{|\mathcal{S}_{j_1}| + |\mathcal{S}_{j_2}|}.
\]

Both variants regularize the initial graph through set unions, yielding weights in $[0, 1]$ and thus a more comparable scale across function pairs than the raw co-occurrence count.
\subsection{LLM-based agents}
\label{app:skill}

We manually curate a unified instruction set for semantic map construction and evaluate two LLM-based agents, Qwen and Codex. For Qwen, the instructions and a case-specific input are provided together, and the model produces the semantic map end-to-end in a single pass. For Codex, the instructions are first given to Codex to generate a reusable semantic-map-construction skill. For each case, we then provide the case data and explicitly invoke the learned skill, so that Codex automatically generates the corresponding semantic map.
\subsubsection{Instruction}
We designed a structured prompt that instructs the model to act as a typological linguist and generate a SMM from multilingual form-function tables. The prompt explicitly requires the construction of a semantic association graph where nodes represent functions and edges denote co-expression relationships, subject to both the connectivity constraint (ensuring each form’s functions induce a connected subgraph) and the minimality constraint (minimizing edges and avoiding cycles). Finally, the model is directed to output the resulting semantic map as an adjacency matrix.
\subsubsection{Qwen}


When we asked the Qwen agent to construct semantic maps directly from the task prompt, we observed two sources of variability. First, the agent often developed ad hoc procedures that were not explicitly specified in the prompt. Second, the procedure varied across runs, resulting in inconsistent semantic maps. Among all cases, the most frequently observed strategy was a coverage-based greedy algorithm, which is neither efficient nor well-suited for constructing semantic maps.

\begin{algorithm}[t]
\caption{Minimum-Edge Semantic Map Construction}
\label{alg:codex}
\begin{algorithmic}[2]
\Require Function sets associated with forms $\{S_r\}$
\Ensure Minimum-edge semantic map $E^{*}$

\State $E \leftarrow
\{\{u,v\}:u,v\in S_r
\text{ for some }r\}$
\State $\mathcal{C}\leftarrow
\Call{BuildCutConstraints}{\{S_r\},E}$
\State Remove redundant constraints from $\mathcal{C}$
\State Compute normalized and raw co-occurrence scores for each $e\in E$

\State $E^{*}\leftarrow
\Call{GreedyUpperBound}{E,\mathcal{C}}$
\State \Call{Search}{$\varnothing,\mathcal{C}$}
\State \Return $E^{*}$

\Procedure{Search}{$E_{\mathrm{cur}},\mathcal{C}_{\mathrm{rem}}$}
    \If{$\mathcal{C}_{\mathrm{rem}}=\varnothing$}
        \If{$|E_{\mathrm{cur}}|<|E^{*}|$}
            \State $E^{*}\leftarrow E_{\mathrm{cur}}$
        \EndIf
        \State \Return
    \EndIf

    \State $L\leftarrow
    \Call{PruningLowerBound}{\mathcal{C}_{\mathrm{rem}}}$
    \If{$|E_{\mathrm{cur}}|+L\geq|E^{*}|$}
        \State \Return
    \EndIf

    \State Select $C^{*}\in\mathcal{C}_{\mathrm{rem}}$
    with the fewest candidate edges
    \State Sort $e\in C^{*}$ by coverage, normalized score,
    raw score, and edge index

    \For{each ordered edge $e\in C^{*}$}
        \State \Call{Search}
        {$E_{\mathrm{cur}}\cup\{e\},
        \{C\in\mathcal{C}_{\mathrm{rem}}:e\notin C\}$}
    \EndFor
\EndProcedure
\end{algorithmic}
\end{algorithm}

\subsubsection{Codex}
To establish a stable and reproducible baseline, semantic maps are constructed using Codex’s reusable semantic-map skill. For each form, the skill identifies its functions and converts the connectivity requirement into cut constraints. Each non-trivial partition of the active functions defines a constraint containing the candidate edges that cross the partition, and a valid map must select at least one edge from every constraint. The problem is therefore formulated as a minimum-cardinality hitting-set problem~\cite{karp1972reducibility}.

The algorithm first obtains a feasible solution through a greedy strategy to initialize an upper bound, and then performs branch-and-bound search. It selects the unresolved constraint with the fewest candidate edges and orders candidate edges by their coverage of unresolved constraints. Ties are broken by normalized and then raw co-occurrence frequency. These statistics affect only the search order and do not alter the minimum-edge objective. The complete procedure is summarized in Algorithm~\ref{alg:codex}. If exhaustive search finishes within the prescribed time limit, the returned map is certified as globally optimal; otherwise, the best feasible solution found so far is returned and marked as unproven.

\subsection{Dataset-specific F1 Comparison}
\label{app:case_f1}

\begin{figure}[t]
    \centering
    \includegraphics[width=1.0\linewidth]{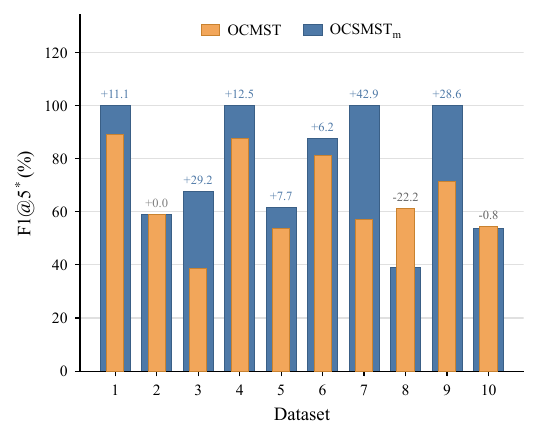}
    \caption{Maximum F1 score among the top 5 candidates for OCMST and our method OCSMST\_m.}
    \label{fig:f1_by_case}
\end{figure}

Figure~\ref{fig:f1_by_case} presents dataset-specific comparisons between OCMST and OCSMST\_m. Performance improves in 7 out of 10 cases, with four datasets (Datasets 1, 4, 7, and 9) reaching an F1 of 100\%. Notably, on Dataset 1, the SUPPLEMENT adverbs case study in OCMST~\cite{liu2025top}, our method generates an SMM that exactly matches human experts, demonstrating its competitiveness.

\subsection{Form-function Table for Dataset 7}
\label{app:city_features}

\begin{table}[htbp]
    \centering
    \small
    \setlength{\tabcolsep}{5pt}
    \renewcommand{\arraystretch}{1.1}
    \begin{tabular}{l cccccccc}
    \toprule
    \textbf{Dialect} & \textbf{E} & \textbf{P} & \textbf{U} & \textbf{H} & \textbf{F} & \textbf{W} & \textbf{Fm} & \textbf{A} \\
    \midrule
    Yinchuan   & 1 & 1 & 1 & 1 & 0 & 0 & 0 & 0 \\
    Taiyuan    & 1 & 1 & 1 & 1 & 0 & 0 & 0 & 0 \\
    Wanrong    & 1 & 1 & 0 & 1 & 0 & 0 & 0 & 0 \\
    Yangzhou   & 1 & 1 & 0 & 1 & 1 & 0 & 1 & 0 \\
    Xuzhou     & 1 & 1 & 0 & 0 & 0 & 1 & 0 & 0 \\
    Xi'an      & 1 & 1 & 0 & 1 & 1 & 0 & 0 & 0 \\
    Leizhou    & 1 & 1 & 1 & 1 & 0 & 1 & 0 & 0 \\
    Pingxiang  & 1 & 1 & 1 & 1 & 0 & 1 & 0 & 0 \\
    Meixian    & 1 & 1 & 1 & 1 & 0 & 1 & 1 & 0 \\
    Xiamen     & 1 & 1 & 1 & 1 & 0 & 1 & 1 & 0 \\
    Jian'ou    & 1 & 1 & 1 & 1 & 0 & 1 & 1 & 1 \\
    Haikou     & 1 & 1 & 1 & 1 & 0 & 0 & 1 & 0 \\
    Jixi       & 1 & 1 & 1 & 1 & 0 & 0 & 1 & 0 \\
    Muping     & 1 & 1 & 1 & 1 & 0 & 0 & 1 & 1 \\
    Guangzhou  & 1 & 1 & 1 & 1 & 0 & 0 & 1 & 1 \\
    Nanjing    & 1 & 1 & 1 & 1 & 0 & 0 & 0 & 1 \\
    Xinzhou    & 1 & 1 & 1 & 1 & 1 & 0 & 0 & 1 \\
    \bottomrule
    \end{tabular}
\caption{Form-function table for Dataset 7, whose domain is MAKE (Chinese \textit{zuò}). Column abbreviations: E = Engage, P = Produce, U = Undertake, H = Hold, F = Feign, W = Write, Fm = Form, A = Act As.}
    \label{tab:city-features}
\end{table}

Table~\ref{tab:city-features} shows the form-function table for Dataset 7 on the semantic domain of MAKE (Chinese \textit{zuò}). The languages are Chinese dialects from different cities, and the function abbreviations follow the header.
\label{sec:appendix}

\end{document}